# Generalized Evolutionary Algorithm based on Tsallis Statistics

Ambedkar Dukkipati,[*] M. Narasimha Murty,[†] and Shalabh Bhatnagar[‡]
*Department of Computer Science and Automation Indian Institute of Science, Bangalore 560012, India*
(Dated: February 1, 2018)

Generalized evolutionary algorithm based on Tsallis canonical distribution is proposed. The algorithm uses Tsallis generalized canonical distribution to weigh the configurations for 'selection' instead of Gibbs-Boltzmann distribution. Our simulation results show that for an appropriate choice of non-extensive index that is offered by Tsallis statistics, evolutionary algorithms based on this generalization outperform algorithms based on Gibbs-Boltzmann distribution.

PACS numbers: Valid PACS appear here

## I. INTRODUCTION

Evolutionary computation comprises techniques for obtaining near-optimal solutions of hard optimization problems in physics [1, 2] and engineering [3]. These methods are based loosely on ideas from biological evolution and are similar to simulated annealing, except that instead of exploring the search space with a single point at each instant, it deals with a population – a multi-subset of search space – in order to avoid getting trapped in local optima during the process of optimization. Though evolutionary algorithms are not known generally as Monte Carlo methods, recently these algorithms have been analyzed in Monte Carlo framework [e.g. 4, 5, 6].

A typical evolutionary algorithm is a two step process: *selection* and *variation*. Selection comprises replicating an individual in the population based on probabilities (these are called selection probabilities) assigned to the individuals in the population on the basis of a "fitness" measure defined by the objective function. Stochastic perturbation of individuals while replicating is called variation.

Selection is a central concept in evolutionary algorithms. There are several selection mechanisms in evolutionary algorithms, among which Boltzmann selection has an important place because of the deep connection between the behavior of complex systems in thermal equilibrium at finite temperature and multivariate optimization [7]. In these systems, each configuration is weighted by its Gibbs-Boltzmann probability factor $e^{-E/T}$, where $E$ is the energy of the configuration and $T$ is the temperature. Finding the low-temperature state of a system when the energy can be computed amounts to solving an optimization problem. This connection has been used to devise the simulated annealing algorithm [8]. Similarly for evolutionary algorithms in the selection process where one would select "better" configurations, one can use the same technique to weigh the individuals i.e., using Gibbs-Boltzmann factor. This is called Boltzmann selection. which is nothing but defining selection probabilities in the form of Gibbs-Boltzmann canonical distribution.

Recently Tsallis and Stariolo [9] proposed generalized simulated annealing based on Tsallis statistics [10]. This method is shown to be faster than both classical simulated annealing and the fast simulated annealing methods [10, 11]. This algorithm has been used successfully in many applications in Physics and Chemistry [12, 13, 14, 15, 16].

This is the motivation for us to incorporate Tsallis canonical probability distribution for selection in evolutionary algorithms instead of Gibbs-Boltzmann distribution. To our knowledge this is the first attempt to use Tsallis probabilities in evolutionary computation and test the novelty of this technique using simulations.

The outline of the paper is as follows. In § II, we review evolutionary algorithms based on Gibbs-Boltzmann distribution and in § III we present our algorithm based on Tsallis generalized distribution. We present simulation results in § IV.

## II. EVOLUTIONARY ALGORITHMS BASED ON GIBBS-BOLTZMANN DISTRIBUTION

Let $\Omega$ be the search space i.e. space of all configurations of an optimization problem and let $E : \Omega \to \mathbb{R}$ be the objective function – following statistical mechanics terminology [e.g. 7, 17] we refer to this function as energy (in evolutionary computation terminology this is called as fitness function) – where the objective is to find a configuration with lowest energy. Let $P = \{\omega_k\}_{k=1}^n$ denote a population which is a multi-subset of $\Omega$. Here we assume that the size of population at any time is finite and need not be a constant.

The general structure of evolutionary algorithm is shown in the FIG 1; for further details refer to [18, 19]. In the first step, population $P(0)$ is initialized with random configurations. At each time step $t$ population undergoes the following procedure.

$$P(t) \xrightarrow{\text{selection}} P'(t) \xrightarrow{\text{variation}} P(t+1)$$

Variation is nothing but stochastically perturbing the individuals in the population. Various methods in evolutionary algorithms follow different approaches. For example in genetic algorithms, where configurations are represented as binary strings, operators such as mutation and crossover are used; for details see [3].

[*]Electronic address: ambedkar@csa.iisc.ernet.in
[†]Electronic address: mnm@csa.iisc.ernet.in
[‡]Electronic address: shalabh@csa.iisc.ernet.in

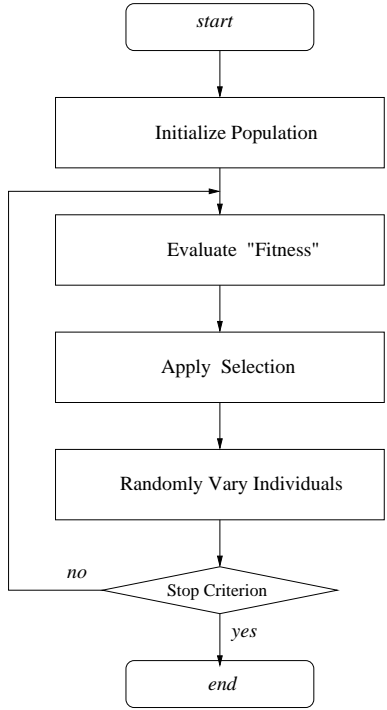

FIG. 1: Structure of evolutionary algorithms

Selection is the mechanism, where the "good" configurations are replicated based on their selection probabilities [20]. For a population $P(t) = \{\omega_k\}_{k=1}^n$ with the corresponding energy values $\{E_k\}_{k=1}^n$, selection probabilities are defined as,

$$p_t(\omega_k) = \text{Prob}(\omega_k \in P'(t)|\omega_k \in P(t)) \ \forall i = 1 \ldots n \ ,$$

and $\{p_t(\omega_k)\}_{k=1}^n$ satisfies the condition:

$$\sum_{k=1}^n p_t(\omega_k) = 1 \ . \qquad (1)$$

According to Boltzmann selection, selection probabilities are defined as

$$p_t(\omega_k) = \frac{e^{-\beta_t E_k}}{\sum_{j=1}^n e^{-\beta_t E_j}} \ , \qquad (2)$$

where $\beta_t$ is the *inverse temperature* at time $t$ and $\{\beta_t : t = 0, 1 \ldots\}$ is an annealing schedule. The strength of selection is controlled by the parameter $\beta$. A higher value of $\beta$ (low temperature) gives a stronger selection, and a lower value of $\beta$ gives a weaker selection [20].

Boltzmann selection gives faster convergence, but without good annealing schedule for $\gamma$, it might lead to premature convergence. This problem is well known from simulated annealing [21], but not very well studied in evolutionary algorithms. Recently Mahnig and Mühlenbein [22], Dukkipati et al. [23] proposed annealing schedules for evolutionary algorithms based on Boltzmann selection.

## III. GENERALIZED EVOLUTIONARY ALGORITHM BASED ON TSALLIS PROBABILITIES

Tsallis [10] proposed a generalization of the celebrated Boltzmann-Gibbs entropy measure (Shannon entropy in information theoretic terms) which is defined as

$$S_q(p) = k \frac{1 - \sum_k p_k{}^q}{q-1} \quad (q \in \mathbb{R}) \ , \qquad (3)$$

where $p = \{p_k\}$ is a probability distribution, $q$ is called non-extensive index and $k$ is a conventional positive constant (we adopt $k=1$ for simplicity). We have

$$\lim_{q \to 1} S_q(p) = -\sum_k p_k \ln p_k = S_1(p) \ , \qquad (4)$$

which is Shannon's entropy, i.e., in the $q \to 1$ limit, $S_q$ recovers the Shannon entropy. Maximizing the Tsallis entropy $S_q$ with the constraints

$$\sum_k p_k = 1 \text{ and } \sum_k p_k{}^q E_k = \text{const}, \qquad (5)$$

where $\{E_k\}$ is the energy spectrum. The generalized probability distribution is found to be [10],

$$p_k = \frac{[1-(1-q)\beta E_k]^{\frac{1}{1-q}}}{Z_q} \ , \qquad (6)$$

with $Z_q$ as the partition function,

$$Z_q = \sum_k [1-(1-q)\beta E_k]^{\frac{1}{1-q}} , \qquad (7)$$

where $\beta = \frac{1}{T}$ is a Lagrange parameter called inverse temperature. This distribution is Tsallis generalized canonical distribution [24]. Note that

$$\lim_{q \to 1} \frac{[1-(1-q)\beta E_k]^{\frac{1}{1-q}}}{Z_q} = \frac{e^{-\beta E_k}}{Z_1} \ , \qquad (8)$$

i.e., Tsallis distribution goes to the Gibbs-Boltzmann distribution when $q$ tends to 1.

Inspired by these general statistics, we propose a new selection scheme for evolutionary algorithms based on Tsallis generalized canonical distribution. For a population $P(t) = \{\omega_k\}_{k=1}^n$ with corresponding energies $\{E_k\}_{k=1}^n$ we define selection probabilities as

$$p_t(\omega_k) = \frac{[1-(1-q)\beta_t E_k]^{\frac{1}{1-q}}}{Z_q} \ \forall k = 1, \ldots n \ . \qquad (9)$$

where $\{\beta_t : t = 1, 2, \ldots\}$ is annealing schedule. We refer selection scheme based on Tsallis distribution as Tsallis selection and evolutionary algorithm with Tsallis selection as generalized evolutionary algorithm.

In this paper we use the annealing schedule derived by Dukkipati et al. [23] which is called Cauchy annealing schedule for evolutionary algorithms. According to this annealing schedule $\beta_t$ should be a non-decreasing Cauchy sequence for faster convergence. In [23] the non-decreasing Cauchy sequence is chosen as

$$\beta_t = \beta_0 \sum_{i=1}^{t} \frac{1}{i^\alpha} \quad t = 1, 2, \ldots \quad , \quad (10)$$

where $\beta_0$ is any constant, $\alpha > 1$ and novelty of this annealing schedule had been demonstrated using simulations. Similar to the practice in generalized simulated annealing [e.g 16], in our algorithms $q$ tends towards 1 as temperature decreases during annealing.

The generalized evolutionary algorithm based on Tsallis statistics is listed in FIG. 2.

---

**Algorithm 1** Generalized Evolutionary algorithm

$P(0) \leftarrow$ Initialize with configurations from search space randomly
Initialize $\beta$ and $q$
**for** $t = 1$ to $T$ **do**
  **for all** $\omega \in P(t)$ **do**
    (*Selection*)
    Calculate
    $$p(\omega) = \frac{[1 - (1-q)\beta E(\omega)]^{\frac{1}{1-q}}}{Z_q}$$
    Copy $\omega$ into $P'(t)$ with probability $p(\omega)$ with replacement
  **end for**
  **for all** $\omega \in P'(t)$ **do**
    (*Variation*)
    Perform variation with specific probability
  **end for**
  Update $\beta$ according to annealing schedule
  Update $q$ according to its schedule
  $P(t+1) \leftarrow P'(t)$
**end for**

---

FIG. 2: Generalized Evolutionary Algorithm based on Tsallis statistics to optimize the energy function $E(\omega)$.

## IV. SIMULATION RESULTS

We discuss the simulations conducted to study generalized evolutionary algorithm based on Tsallis statistics proposed in this paper. We compare performance of evolutionary algorithms with three selection mechanisms viz., proportionate selection (where selection probabilities of configurations are inversely proportional to their energies [3]), Boltzmann selection and Tsallis selection. For comparison purposes we study multi-variable function optimization in the framework of genetic algorithms. Specifically, we use the following bench mark test functions [25], where the aim is to find the configuration with lowest functional value:

- Ackley's function:
  $E_1(\vec{x}) = -20 \exp(-0.2\sqrt{\frac{1}{l}\sum_{i=1}^{l} x_i^2})$
  $- \exp(\frac{1}{l}\sum_{i=1}^{l} \cos(2\pi x_i)) + 20 + e$,
  where $-30 \leq x_i \leq 30$

- Rastrigin's function:
  $E_2(\vec{x}) = lA + \sum_{i=1}^{l} x_i^2 - A\cos(2\pi x_i)$,
  where $A = 10$ ; $-5.12 \leq x_i \leq 5.12$

- Griewangk's function:
  $E_3(\vec{x}) = \sum_{i=1}^{l} \frac{x_i^2}{4000} - \prod_{i=1}^{l} \cos(\frac{x_i}{\sqrt{i}}) + 1$,
  where $-600 \leq x_i \leq 600$

Parameters for the algorithms were set to compare performance of these algorithms in identical conditions. Each $x_i$ is encoded with 5 bits and $l = 15$ i.e., search space is of size $2^{75}$. Population size is $n = 350$. For all the experiments, probability of uniform crossover is 0.8 and probability of mutation is below 0.1. We limited each algorithm for 100 iterations and we have given the plots for behavior of the process when averaged over 20 runs.

As we mentioned earlier, for Boltzmann selection we have used Cauchy annealing schedule (see (10)) and $\beta_0 = 200$ and $\alpha = 1.01$. For Tsallis selection too we have used the same annealing schedule as Boltzmann selection with identical parameters. In our preliminary simulations $q$ was kept constant and tested with various values. Then we adopted a strategy from generalized simulated annealing where one would choose an initial value of $q_0$ and decrease linearly to the value 1. This schedule of $q$ gave better performance than keeping it constant. We reported results with various values of $q_0$.

From various simulations we observed that when the problem size is small (for example smaller values of $l$) all the selection mechanisms perform equally well. Boltzmann selection is effective when we increase the problem size. The novelty of Cauchy annealing schedule for Boltzmann selection is shown in [23]. For Tsallis selection we performed simulations with various values of $q_0$. FIG. 3 shows the performance for Ackley function for $q_0 = 3, 2, 1.5$ and $1.01$ from which one could conclude that the choice of $q_0$ is very important for evolutionary algorithm with Tsallis selection which varies with problem at the hand.

FIG. 4, 5 and 6 show the comparisons of evolutionary algorithms based on Tsallis selection, Boltzmann selection and proportionate selection for different functions.

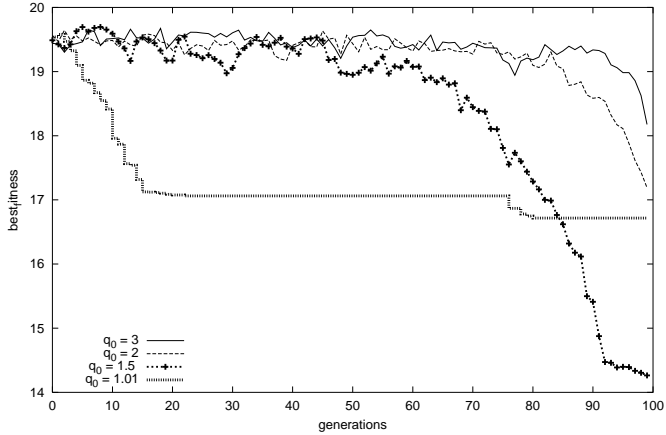

FIG. 3: Performance of evolutionary algorithm with Tsallis selection for various values of $q_0$ for the test function Ackley

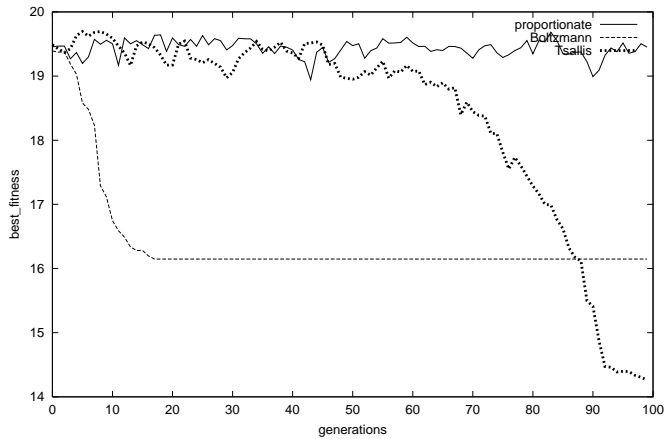

FIG. 4: Ackley: $q_0 = 1.5$

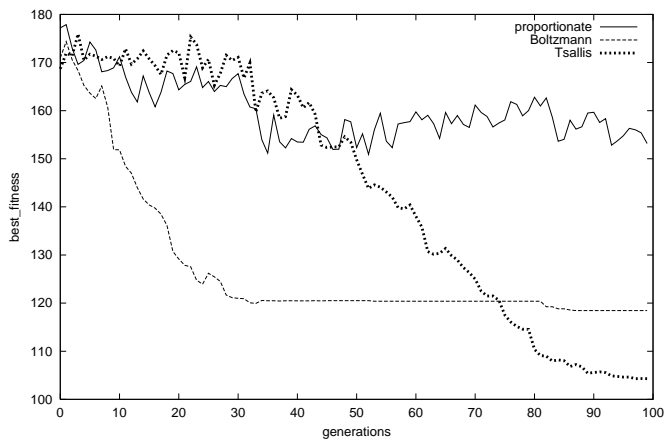

FIG. 5: Rastrigin: $q_0 = 2$

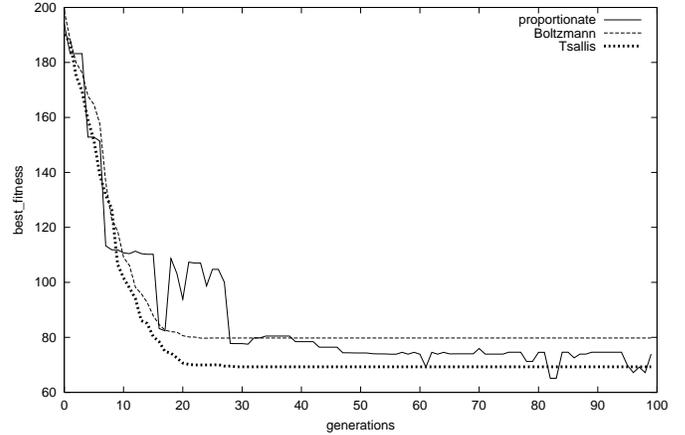

FIG. 6: Griewangk: $q_0 = 1.01$

We have reported only the best behavior for various values of $q_0$. From these simulation results we conclude that the evolutionary algorithm based on Tsallis canonical distribution with appropriate value of $q_0$ outperform algorithms based on Boltzmann and proportionate selection.

## V. CONCLUSION

Inspired by generalization of simulated annealing reported by Tsallis and Stariolo [9], in this paper we proposed a generalized evolutionary algorithm based on Tsallis statistics. The algorithm uses Tsallis canonical probability distribution instead of Gibbs-Boltzmann distribution.

We tested our algorithm on bench-mark test functions. We found that with an appropriate choice of non-extensive parameter ($q$), evolutionary algorithm based on Tsallis statistics outperforms evolutionary algorithms based on Gibbs-Boltzmann distribution. We believe the Tsallis canonical distribution is a powerful technique for selection mechanism in evolutionary algorithms. Evolutionary algorithms have been analyzed in the past using statistical mechanics [17, 26, 27, 28], therefore analysis of the same using non-extensive statistical mechanics [29] would be very welcome.

### Acknowledgments

Research work reported here is supported in part by AOARD Grant F62562-03-P-0318 and by DST Grant SR/S3/EE/43/2002-SERC-Engg.